\pdfoutput=1

\documentclass[11pt]{article}

\usepackage[final]{acl}

\usepackage{times}
\usepackage{latexsym}

\usepackage[T1]{fontenc}

\usepackage[utf8]{inputenc}

\usepackage{microtype}

\usepackage{inconsolata}

\usepackage{graphicx}
\usepackage{subcaption}

\usepackage{amsmath}
\usepackage{booktabs}
\usepackage{multirow}
\usepackage{amssymb}
\usepackage{color}

\usepackage{xspace}
\newcommand\ourscore{DBG\xspace}
\title{Beyond the Surface: Measuring Self-Preference in LLM Judgments}

\author{
 \textbf{Zhi-Yuan Chen\textsuperscript{1}},
 \textbf{Hao Wang\textsuperscript{2}},
 \textbf{Xinyu Zhang\textsuperscript{2}},
 \textbf{Enrui Hu\textsuperscript{2}},
 \textbf{Yankai Lin\textsuperscript{1,3}\thanks{Corresponding author}}
\\
\\
 \textsuperscript{1}Gaoling School of Artificial Intelligence, Renmin University of China, \\
 \textsuperscript{2}Huawei Poisson Lab, \\
 \textsuperscript{3}Beijing Key Laboratory of Research on Large Models and Intelligent Governance
\\
 \small{
   \textbf{Correspondence:} \href{zhiyuan.chen2001@gmail.com}{zhiyuan.chen2001@gmail.com} \quad \href{yankailin@ruc.edu.cn}{yankailin@ruc.edu.cn}
 }
}

\begin{document}
\maketitle
\begin{abstract}

Recent studies show that large language models (LLMs) exhibit self-preference bias when serving as judges, meaning they tend to favor their own responses over those generated by other models. Existing methods typically measure this bias by calculating the difference between the scores a judge model assigns to its own responses and those it assigns to responses from other models. However, this approach conflates self-preference bias with response quality, as higher-quality responses from the judge model may also lead to positive score differences, even in the absence of bias. To address this issue, we introduce gold judgments as proxies for the actual quality of responses and propose the DBG score, which measures self-preference bias as the difference between the scores assigned by the judge model to its own responses and the corresponding gold judgments. Since gold judgments reflect true response quality, the DBG score mitigates the confounding effect of response quality on bias measurement. Using the DBG score, we conduct comprehensive experiments to assess self-preference bias across LLMs of varying versions, sizes, and reasoning abilities. Additionally, we investigate two factors that influence and help alleviate self-preference bias: response text style and the post-training data of judge models. Finally, we explore potential underlying mechanisms of self-preference bias from an attention-based perspective. Our code and data are available at \url{https://github.com/zhiyuanc2001/self-preference}.


\end{abstract}

\begin{figure}[!t]
    \centering
    \includegraphics[width=1.0\linewidth]{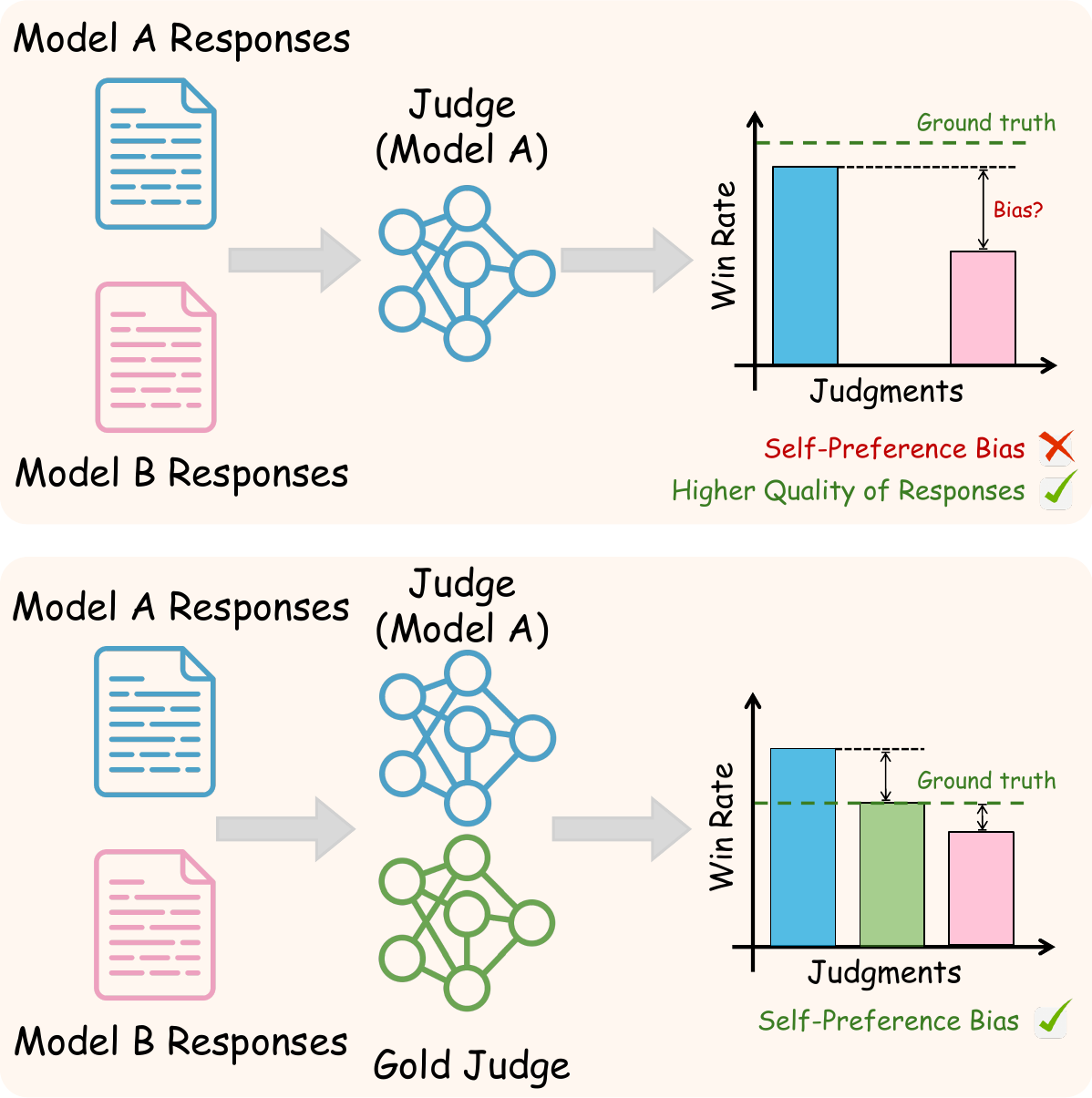}
    \caption{Current methods (Top) measure the self-preference bias of the judge model by comparing the scores (such as win rate) that the judge model assigns to its own responses with those assigned to other models' responses. However, these methods overlook the impact of the intrinsic quality of the responses on the scores provided by the judge model. Our approach (Bottom) introduces gold judgments as proxies for the ground truth scores of responses. By comparing the scores that the judge model gives to its own responses with gold judgments, our method can provide a more reliable assessment of the self-preference bias.}
    \label{fig:flowchart}
    \vspace{-12pt}
\end{figure}

\section{Introduction}

Comprehensive evaluation of large language models (LLMs) has become a central and evolving research challenge in recent years. As tasks become increasingly diverse and complex, traditional rule-based (e.g., BLEU~\citep{papineni2002bleu}) and human-based evaluation approaches encounter significant limitations. Rule-based approaches often lack flexibility in open-ended tasks, while human-based approaches are prohibitively expensive and time-consuming~\citep{hendrycks2021measuring, chiang2024chatbot}. Recently, LLMs as judges is proposed as a valuable complement to both rule-based and human-based evaluation approaches~\citep{zheng2023judging, li2024generation}. By leveraging their extensive world knowledge and reasoning abilities, LLMs show a high degree of alignment with human judgments and offer a convenient, cost-effective alternative to human-based evaluation~\citep{zheng2023judging, zhu2023judgelm}. While LLMs are widely employed as judges, empirical evidence indicates that they are susceptible to \textit{self-preference} bias, which refers to the tendency of LLMs to assign higher scores to their own responses compared to those generated by other models~\citep{liu2023llms, wataoka2024self, chen2025llm}. Self-preference bias leads LLMs to produce inaccurate judgment results, undermining their reliability as judges.

To measuring the self-preference bias of a judge model, existing work typically uses the difference between the scores the judge model assigns to its own responses and those it assigns to other models’ responses as the bias indicator (as shown in Figure~\ref{fig:flowchart}). However, this approach conflates response quality with the judge model’s self-preference bias~\citep{chen2025llm}, potentially leading to inaccurate assessments. Specifically, if the judge model produces high-quality responses, it becomes ambiguous whether the high scores it assigns to its own responses are due to their actual quality or due to self-preference bias. 

To address this issue, we introduce gold judgments and use them as proxies for the ground truth scores of responses. We then propose the \ourscore score, which \textbf{measures the degree of a model’s self-preference bias as the difference between the scores it assigns to its own responses and the corresponding gold judgments}. Subtracting the gold judgments from the scores assigned by the judge model helps isolate self-preference bias and reduces the confounding effect of response quality on the bias measurement (Section~\ref{sec:evaluation}). 
To obtain gold judgments in this setting, we aggregate evaluation results from multiple strong LLMs. By leveraging the consensus among these models, the gold judgments offer a reliable estimate of the true response scores. 

Based on the \ourscore score, we conduct comprehensive experiments to investigate self-preference bias across judge models of different versions, sizes, and reasoning abilities. For model versions, we consider both pre-trained and post-trained variants of LLMs. We observe that both pre-trained and post-trained models exhibit self-preference bias to some extent. Interestingly, although post-trained models undergo additional training based on their pre-trained counterparts, they do not necessarily exhibit a more severe degree of self-preference bias. Regarding model size, we examine models ranging from 0.5B to 72B and find that larger models tend to exhibit less self-preference bias than their smaller counterparts. For reasoning ability, we study large reasoning models (LRMs)~\citep{jaech2024openai, guo2025deepseek} and find that LRMs also display self-preference bias when serving as judges. Notably, the severity of this bias is not necessarily less pronounced than that observed in LLMs.

Furthermore, to investigate the factors that influence and potentially mitigate self-preference bias in models, we explore two key aspects: response text style~\citep{ostheimer2023text} and post-training data. Empirical experiments show that aligning the response styles of different models to a unified style helps alleviate self-preference bias. In addition, training two different types of models on the same dataset encourages a reduction in self-preference bias in both models. Attention-level analysis reveals that, during judgment, models naturally tend to assign higher attention scores to their own responses compared to those generated by the other model, which may partly explain the presence of self-preference bias.

In summary, our contributions are as follows. (1) We propose the \ourscore score to enable more accurate and reliable evaluation of self-preference bias. (2) We conduct comprehensive experiments to measure the self-preference bias of models with varying versions, sizes, and reasoning abilities. (3) We analyze the impact of response text style and post-training data on the self-preference bias of LLMs and offer an attention-based explanation of its potential causes.

\section{The \ourscore Score: Measuring Self-Preference in Judge Models}\label{sec:evaluation}

Self-preference, also known as self-enhancement, refers to the tendency of an LLM to favor its own generated responses when making judgments~\citep{zheng2023judging}. 
Formally, let $A$ and $B$ denote two different LLMs, and let $r_A$ and $r_B$ represent the responses generated by $A$ and $B$, respectively, in response to the same instruction $x$. For simplicity, we focus our analysis on the scenario where model $A$ serves as the judge.

Let $S_A(r)$ denote the score assigned by judge $A$ to response $r$. Following the Bradley-Terry model~\citep{bradley1952rank}, the probability that judge $A$ prefers $r_A$ over $r_B$ is given by:
\begin{equation}
    \mathbb{P}(r_A \succ r_B \mid x) = \sigma(S_A(r_A) - S_A(r_B)), \nonumber
\end{equation}
where $\sigma$ is the sigmoid function. Assume that each response $r$ has an underlying true quality $Q(r)$, and that judge $A$ has an inherent bias $b_A(r)$ toward response $r$. We approximate the score as: $S_A(r) \approx Q(r) + b_A(r)$ and obtain
\begin{equation}
    \mathbb{P}(r_A \succ r_B \mid x) = \sigma(\delta + b_A), \nonumber
\end{equation}
where $\delta = Q(r_A) - Q(r_B)$ captures the quality gap between the two responses, and $b_A = b_A(r_A) - b_A(r_B)$ reflects the bias of judge $A$. In the self-preference bias case, we assume that the judge exhibits bias only toward its own response, such that $b_A(r_B) = 0$ and $b_A = b_A(r_A) > 0$. The expected preference probability of judge $A$ choosing its own response $r_A$ over all instructions is
\begin{equation}
    w_A = \mathbb{E}_x[\sigma(\delta + b_A)]. \nonumber
\end{equation}
In contrast, for an unbiased gold judge, the expected preference probability of selecting $r_A$ is
\begin{equation}
    w^* = \mathbb{E}_x[\sigma(\delta)]. \nonumber
\end{equation}

Recent work adopts metrics based on $w_A$ to quantify self-preference bias~\citep{panickssery2024llm, ye2024justice}. However, this approach conflates the quality of the responses with the self-preference bias of the judge model~\citep{chen2025llm}, leading to biased estimations. Specifically, when models $A$ and $B$ correspond to a strong LLM (e.g., GPT-4o~\citep{hurst2024gpt}) and a weaker LLM (e.g., Llama-3.1-8B-Instruct~\citep{grattafiori2024llama}), it becomes ambiguous whether a higher $w_A$ is driven by inherent differences in response quality or by the self-preference bias of the judge model $A$. 

To isolate the self-preference bias of model $A$, we propose using the \textbf{d}ifference between the \textbf{b}iased judge and the \textbf{g}old judge as a metric (referred to as the \textbf{\ourscore} score) for measuring self-preference bias:
\begin{equation}
    \hat{w}_A = \mathbb{E}_x[\sigma(\delta + b_A) - \sigma(\delta)]. \nonumber
\end{equation}
This formulation removes the confounding effect of response quality (captured by $\delta$) and focuses explicitly on the self-preference bias term $b_A$. A \ourscore score greater than zero indicates that the model exhibits self-preference bias, with larger values suggesting a more severe degree of bias. 

When $b_A$ is small, a first-order Taylor approximation yields
\begin{equation}
    \hat{w}_A \approx \mathbb{E}_x[\sigma'(\delta) \cdot b_A]. \nonumber
\end{equation}
Assuming a weak correlation between response quality gaps and self-preference bias of $A$, we have
\begin{equation}
    \hat{w}_A \approx \mathbb{E}_x[\sigma'(\delta)] \cdot \mathbb{E}_x[b_A], \nonumber
\end{equation}
suggesting that $\hat{w}_A$ serves as a linearly scaled estimator of the true bias. Thus, it offers a more accurate and disentangled measure of self-preference than $w_A$.

In practice, we aggregate the judgment results from three strong LLMs to construct the unbiased gold judgment:
\begin{equation}
    \hat{w}^* = \mathbb{E}_{x,k}[\sigma(\delta + b_k)], \nonumber
\end{equation}
where $b_k$ denotes the bias of model $k$ toward $r_A$. Using the Taylor expansion, we obtain:
\begin{equation}
    \hat{w}^* = \mathbb{E}_x[\sigma(\delta)] + \mathbb{E}_{x,k}[\Delta], \nonumber
\end{equation}
where $\Delta$ represents the remainder term. If the bias of each individual model is relatively small or fluctuates around zero, then $\Delta \approx 0$. This indicates that aggregation helps mitigate the bias of any single model and enhances the stability of the evaluation. Additionally, to further validate the reliability of the gold judgments, we conduct a human study, as detailed in Section~\ref{human_study}.

\begin{figure*}[!ht]
  \centering

  \begin{subfigure}{1\textwidth}
    \centering
    \includegraphics[width=1\textwidth]{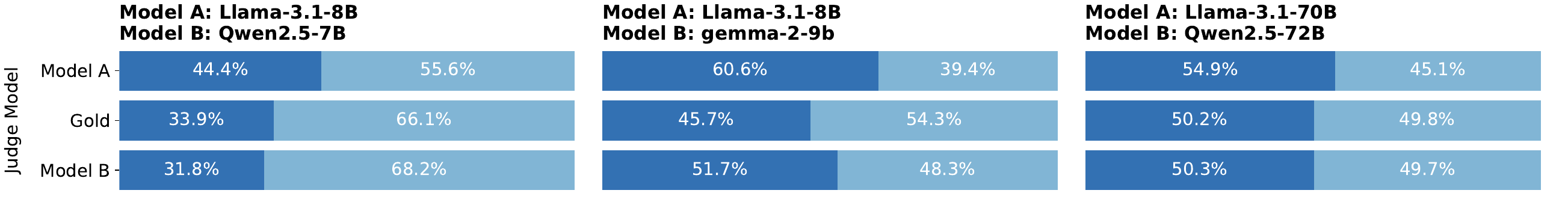}
    \captionsetup{skip=1pt}
    \caption{Pairs of pre-trained models.}
  \end{subfigure}

    \vspace{3pt}

  \begin{subfigure}{1\textwidth}
    \centering
    \includegraphics[width=1\textwidth]{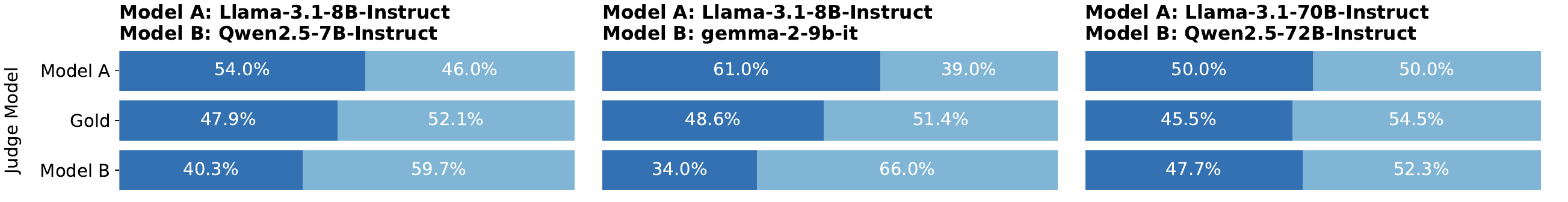}
    \captionsetup{skip=1pt}
    \caption{Pairs of post-trained models.}
  \end{subfigure}

    \vspace{3pt}

  \begin{subfigure}{1\textwidth}
    \centering
    \includegraphics[width=1\textwidth]{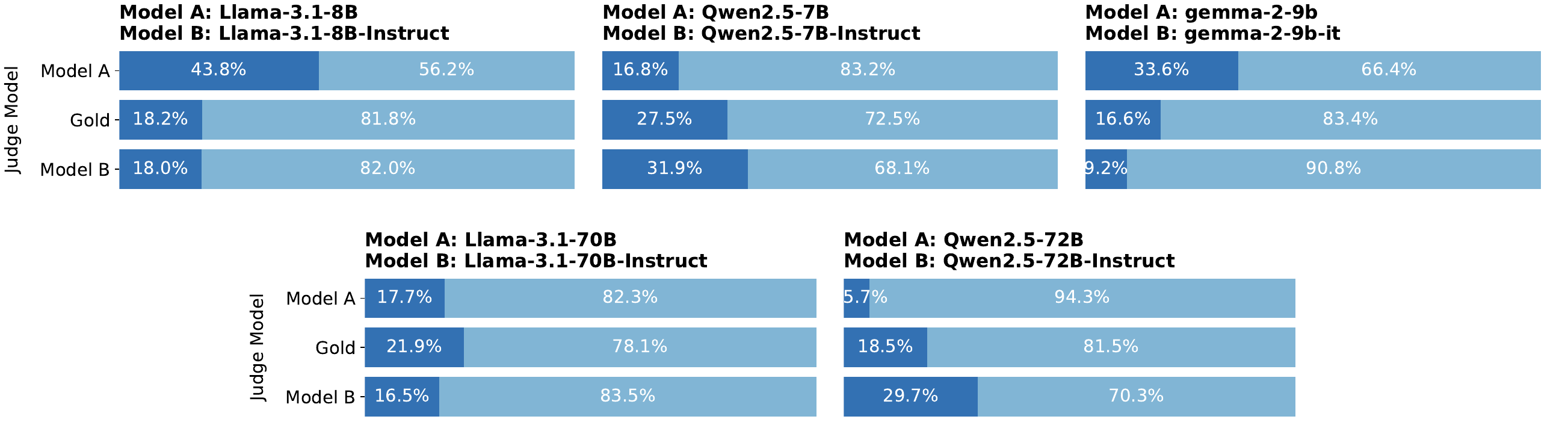}
    \captionsetup{skip=1pt}
    \caption{Pairs of pre-trained and post-trained models.}
  \end{subfigure}

  \vspace{3pt}

  \begin{subfigure}{0.35\textwidth}
    \centering
    \includegraphics[width=\textwidth]{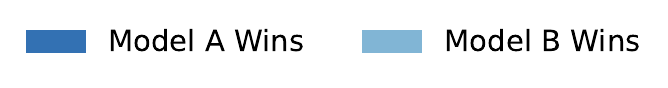}
  \end{subfigure}
    \vspace{-5pt}
  \caption{Judgment results for model pairs of the same size on AlpacaEval.}
    \vspace{-10pt}
  \label{fig:alpaca_main}
\end{figure*}

\section{Experiments}\label{sec:experiments}

\subsection{Experimental Setup}
\paragraph{Models and Datasets.}~We select GPT-4o-mini~\citep{hurst2024gpt}, Gemini-1.5-Flash~\citep{team2024gemini}, and DeepSeek-V3~\citep{liu2024deepseek} as gold judge models due to their strong judging capabilities. To avoid preference leakage, we choose models of different types from the gold judge models to test self-preference bias. Specifically, we select Llama-3.1-8B(-Instruct), Llama-3.1-70B(-Instruct)~\citep{grattafiori2024llama}, Qwen2.5-7B(-Instruct), Qwen2.5-72B(-Instruct)~\citep{yang2024qwen2}, and gemma-2-9B(-it)~\citep{team2024gemma}, where "-Instruct" and "-it" indicate models that have undergone post-training. We also discuss proprietary models in Appendix~\ref{appendix:proprietary_models}.


We conduct our experiments on three widely-used datasets: AlpacaEval~\citep{alpaca_eval}, WMT19 (de-en)~\citep{wmt19translate} and TruthfulQA~\citep{lin2021truthfulqa}. Following prior work on multi-objective alignment~\citep{cui2023ultrafeedback, guo2024controllable}, we evaluate \textbf{helpfulness} on AlpacaEval and WMT19 (de-en), and \textbf{truthfulness} on TruthfulQA.
To facilitate experiments and ensure reliable evaluation, we randomly sample 500 examples from each dataset.

\paragraph{Implementation Details.}~For all models, we set the temperature to $0$ to ensure output determinism and consistency. For pre-trained LLMs, we leverage the in-context learning method~\citep{brown2020language} and prepend two examples to the prompt, enabling them to generate judgments. Given an LLM and two responses, where one response is generated by the LLM itself, we evaluate the two responses using a pairwise comparison approach. Compared to single-response scoring methods, the pairwise comparison approach yields more stable evaluation results~\citep{zheng2023judging}.

We denote each input to the judge model as $(p, r_A, r_B)$, where $p$ is the judge prompt. This prompt instructs the LLM to judge which of $r_A$ and $r_B$ is better and to output only token \texttt{A} or \texttt{B}. We collect and normalize the probabilities corresponding to the output tokens \texttt{A} and \texttt{B}. To mitigate the impact of position bias~\citep{zheng2023judging, ye2024justice} on the evaluation results, we swap the order of the responses and compute the average probability for each response across both positions~\citep{panickssery2024llm}. Finally, we select the response with the highest average probability as the winner and calculate the win rate over all instructions. The consistency between the theoretical analysis and the empirical implementation is discussed in Appendix~\ref{appendix:consistency}. For gold judgments, since some models do not provide output probabilities, we assign a probability of $1.0$ to the output token from gold judge models and $0.0$ to the other token. Then, we select the winner by averaging the probabilities of all three gold judge models. Furthermore, we alleviate the influence of length bias by constraining the maximum length of the responses. The detailed prompts are presented in Appendix~\ref{appendix:prompt}.

\subsection{Main Results}\label{sec:main_results}

To implement the pairwise comparison judge method, we combine two LLMs into a pair and have each LLM judge the responses generated by the two LLMs in the pair. This approach can simultaneously capture the self-preference bias of the two LLMs. LLM pairs are formed based on model version and model size. Experimental results on the AlpacaEval dataset are shown in Figure~\ref{fig:alpaca_main} and Figure~\ref{fig:alpaca_big_small}. Additional experimental results are presented in Appendix~\ref{appendix:more_main_results}. Based on the figures, we observe that:

\begin{figure*}[!ht]
    \centering
    \begin{subfigure}{1\textwidth}
        \centering
        \includegraphics[width=1\textwidth]{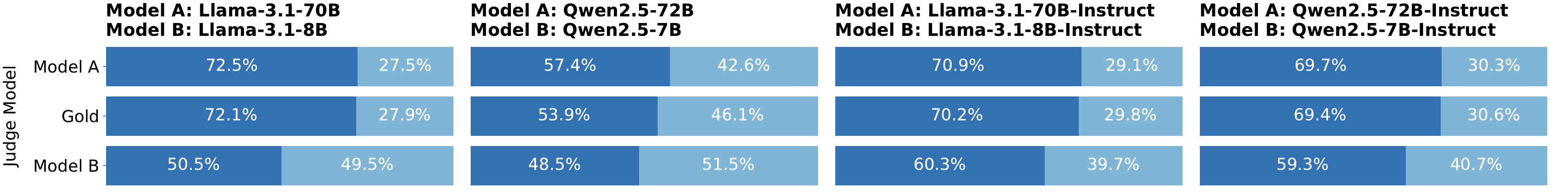}
    \end{subfigure}

    \begin{subfigure}{0.32\textwidth}
        \centering
        \includegraphics[width=\textwidth]{latex/figure/win_rate/main/legend.pdf}
    \end{subfigure}
    \vspace{-5pt}

    \caption{Judgment results for model pairs of different sizes on AlpacaEval.}
    \label{fig:alpaca_big_small}
    \vspace{-10pt}
\end{figure*}

(1) Introducing gold judgments makes the evaluation of self-preference bias more accurate. 
From Figure~\ref{fig:alpaca_main} (b), we observe that when Qwen2.5-72B-Instruct is used as the judge, the win rate score of its responses is $52.3\%$, which is higher than the score obtained when Llama-3.1-70B-Instruct is used as the judge ($50.0\%$), but still falls short of the win rate score given by the gold judgment ($54.5\%$). This suggests that the higher score of Qwen2.5-72B-Instruct may be attributed to the superior quality of its own responses, rather than the self-preference bias. This confirms that introducing gold judgments is necessary to more accurately measure self-preference bias.

(2) Both pre-trained and post-trained models exhibit a certain degree of self-preference bias. Figure~\ref{fig:alpaca_main} (a) shows that when Llama-3.1-8B is paired with Qwen2.5-7B and gemma-2-9B, it assigns higher win rate scores to its own responses than gold judgments do. This indicates that Llama-3.1-8B, when acting as the judge, tends to favor its own responses, resulting in biased scores. Additionally, as shown in Figure~\ref{fig:alpaca_main} (b), we observe that the \ourscore score of Llama-3.1-8B-Instruct is also greater than zero. Larger models, such as Llama-3.1-70B and Llama-3.1-70B-Instruct, exhibit a similar phenomenon. These results suggest that the self-preference bias exists after the pre-training phase and is not solely introduced by the post-training phase.

(3) Post-trained models do not exhibit a more pronounced self-preference bias than pre-trained models. Since post-trained models are further fine-tuned from pre-trained models, an intuitive question arises: does the post-training process intensify the self-preference bias? Figure~\ref{fig:alpaca_main} (c) shows that the self-preference bias in post-trained models is not more severe than in their pre-trained counterparts. In fact, the \ourscore score of Llama-3.1-8B-Instruct is lower than that of Llama-3.1-8B ($0.2\%$ vs. $25.6\%$).

(4) Larger models exhibit less self-preference bias compared to smaller models. As shown in Figure~\ref{fig:alpaca_big_small}, although all models demonstrate self-preference, a noticeable distinction is that the \ourscore scores of larger models are closer to 0. For instance, the \ourscore score of Llama-3.1-70B is $0.4\%$, whereas that of Llama-3.1-8B is $21.6\%$, which is much higher than the score of Llama-3.1-70B. We hypothesize that this may be due to the enhanced instruction-following and judgment capabilities of the larger models, which allow them to assess response quality more fairly and accurately.

\begin{figure}[!htbp]
    \centering
    \includegraphics[width=0.95\linewidth]{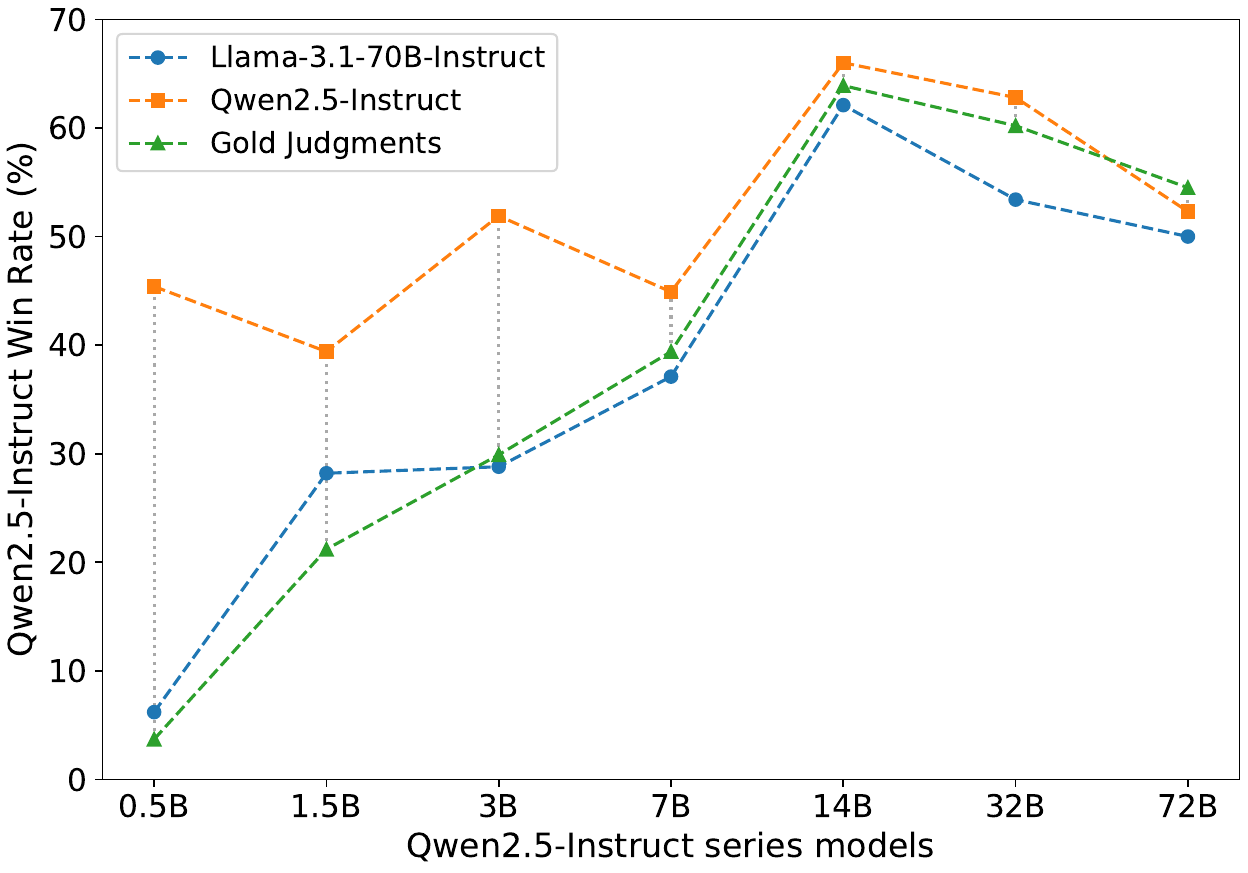}
    \vspace{-5pt}
    \caption{Judgment results for Qwen2.5-Instruct models at different scales.}
    \label{fig:scaling}
    \vspace{-5pt}
\end{figure}

To further investigate how self-preference bias varies with model scale, we conduct experiments using Qwen2.5-Instruct models of different sizes, ranging from 0.5B to 72B. Each model is paired with Llama-3.1-70B-Instruct for judgment. Figure~\ref{fig:scaling} illustrates the win rate of Qwen2.5-Instruct responses under various judge models as the model size increases. As observed in the figure, models larger than 7B exhibit significantly less self-preference bias compared to those of 7B or smaller. For example, the \ourscore score of Qwen2.5-0.5B-Instruct is $41.7\%$. In contrast, the \ourscore score of Qwen2.5-14B-Instruct is only $2.1\%$. This suggests that LLM judging tasks should utilize larger models to obtain more accurate and unbiased judgment results.


\subsection{Alignment Between Gold Judgments and Human Annotations}\label{human_study}

In our experiments, we aggregate the judgment results from three models to serve as gold judgments, which is then used as a reference to measure self-preference bias. To validate the reliability of the gold judgments, we compare it with actual human annotations. Specifically, we randomly sample $100$ instructions from AlpacaEval and obtain the corresponding responses from Llama-3.1-70B-Instruct and Llama-3.1-8B-Instruct. We guide human annotators to compare the responses generated by the two models and determine which response is more helpful. The experimental results are presented in Table~\ref{tab:human_judgment}. From the table, we observe a high degree of consistency between gold judgments and human annotations. On the $100$ samples, the human-annotated win rate for Llama-3.1-70B-Instruct is $63\%$, whereas the gold judgment indicate a win rate of $66\%$. More specifically, we find that human annotations and gold judgment results agree on $74\%$ of the samples. These experimental results validate the reliability and effectiveness of using gold judgments.

\begin{table}[!htbp]
    \small
    \centering
    \setlength{\tabcolsep}{10pt}
    \begin{tabular}{crrr}
        \toprule
        \multirow{2}{*}{Model Pair} & \multicolumn{2}{c}{Judgment} \\
        \cmidrule(lr){2-3}
        & Gold & Human \\
        \midrule
        \ \ Llama-3.1-70B-Instruct & 66.0\% & 63.0\% \\
        Llama-3.1-8B-Instruct & 34.0\% & 37.0\% \\
        \bottomrule
    \end{tabular}
    \caption{Comparison between gold judgments and human annotation results.}
    \label{tab:human_judgment}
    \vspace{-10pt}
\end{table}

\section{Further Analysis}~\label{sec:further_analysis}
In this section, we analyze the self-preference bias exhibited by models of different reasoning abilities. Additionally, we investigate two key factors that influence and help mitigate self-preference: response text style and post-training data. We further explore the underlying mechanisms of self-preference from the perspective of attention. All experiments are conducted on the AlpacaEval dataset.

\subsection{Self-Preference in Reasoning Models}\label{sec:reasoning_models}
To investigate the impact of reasoning ability on model self-preference bias, we test the self-preference bias of DeepSeek(DS)-R1-Distill-Qwen-32B~\citep{guo2025deepseek} and QwQ-32B~\citep{qwq32b}, and compare the results with those of Qwen2.5-32B-Instruct. For LRMs, we remove the reasoning content generated by the models and retain only the final answer for judgment. Since all models are trained on Qwen2.5-32B, this setup mitigates the influence of model size and pre-training process on the results. The experimental results are shown in Table~\ref{tab:reasoning_ability}.

As evidenced in the table, both LRMs exhibit the phenomenon of self-preference bias, as they assign higher win rates to their own responses compared to gold judgments. Notably, although QwQ-32B is capable of generating high-quality responses (with win rate scores from all judge models significantly surpassing those for Llama-3.1-70B-Instruct), it still displays a slight self-preference bias during judgment. Furthermore, we observe that the self-preference bias in reasoning models is not necessarily less significant than the bias found in language models. For instance, the \ourscore score of DS-R1-Distill-Qwen-32B is $4.8\%$, whereas the \ourscore score of Qwen2.5-72B-Instruct is only $2.6\%$. This highlights the importance of addressing judge bias when employing reasoning models as judges in subsequent studies.

\begin{table}[!htbp]
    \footnotesize
    \centering
    \setlength{\tabcolsep}{3pt}
    \begin{tabular}{lrrr}
        \toprule
        \multirow{2}{*}{\centering Model Pair} & \multicolumn{3}{c}{Judge Model} \\
        \cmidrule(lr){2-4}
        & Model A & Gold & Model B \\
        \midrule
        A: Llama-3.1-70B-Instruct & 46.6\% & 39.8\% & 37.2\% \\
        B: Qwen2.5-32B-Instruct & 53.4\% & 60.2\% & 62.8\% \\
        \midrule
        A: Llama-3.1-70B-Instruct & 55.8\% & 51.0\% & 46.2\% \\
        B: DS-R1-Distill-Qwen-32B & 44.2\% & 49.0\% & 53.8\% \\
        \midrule
        A: Llama-3.1-70B-Instruct & 12.4\% & \ 7.6\% & \ 7.0\% \\
        B: QwQ-32B & 87.6\% & 92.4\% & 93.0\% \\
        \bottomrule
    \end{tabular}
    \caption{Self-preference analysis of reasoning models.}
    \label{tab:reasoning_ability}
    \vspace{-10pt}
\end{table}

\begin{figure*}[!ht]
    \centering
    \includegraphics[width=1\linewidth]{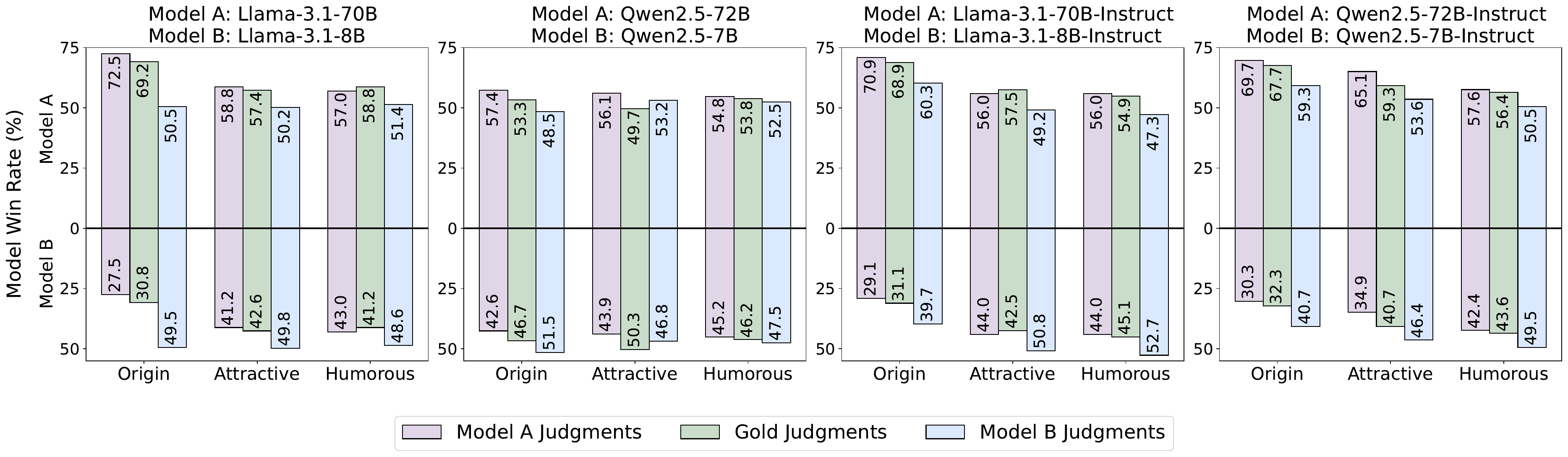}
    \caption{Analysis of response style transfer on model self-preference.}
    \label{fig:text-transfer}
\end{figure*}

\begin{figure*}[!ht]
  \centering
  \begin{subfigure}{1\textwidth}
    \centering
    \includegraphics[width=1\textwidth]{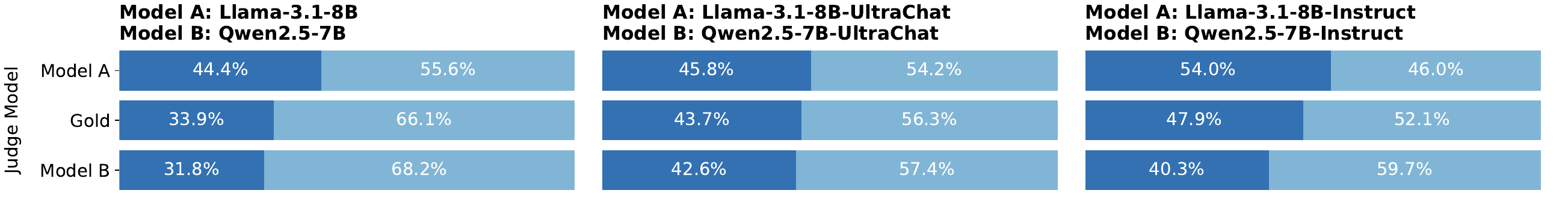}
  \end{subfigure}

  \vspace{-2pt}

  \begin{subfigure}{0.32\textwidth}
    \centering
    \includegraphics[width=\textwidth]{latex/figure/win_rate/main/legend.pdf}
  \end{subfigure}

    \vspace{-10pt}
  \caption{Analysis of post-training data on model self-preference.}
  \label{fig:train_data}
  \vspace{-5pt}
\end{figure*}

\subsection{Impact of Response Style on Self-Preference}\label{sec:style_transfer}
In this section, we investigate whether the superficial linguistic style of LLM-generated responses influences and helps mitigate LLM self-preference. To do so, we modify the response styles and compare the changes in model self-preference bias before and after the modifications. Specifically, for a pair of models, we prompt DeepSeek-V3 to uniformly rewrite the responses of both models into \textbf{attractive} and \textbf{humorous} styles~\citep{ostheimer2023text, mir2019evaluating}. Since DeepSeek-V3 is used to modify the response styles, we exclude it from the gold judge models to mitigate its potential impact on the results. Experimental results are presented in Figure~\ref{fig:text-transfer}. In Appendix~\ref{appendix:content_var}, we provide evidence that our rewriting method minimally affects the semantic content of the responses, thus ensuring that variations in content do not confound the experimental outcomes.

From the figure, we observe that modifying the style of model responses helps alleviate the self-preference bias exhibited by the models when acting as judges. For example, considering the pre-trained models Llama-3.1-70B and Llama-3.1-8B, before style modifications, their \ourscore scores are $3.3\%$ and $18.7\%$, respectively. After rewriting their responses into the attractive style, the scores decrease to $1.4\%$ and $7.2\%$, respectively. Similarly, the post-trained models Qwen2.5-72B-Instruct and Qwen2.5-7B-Instruct exhibit \ourscore scores of $2.0\%$ and $8.4\%$, respectively, before style modifications. After rewriting the responses into the humorous style, the scores decrease to $1.2\%$ and $5.9\%$, respectively. Furthermore, we note that style modifications alone do not entirely eliminate the model self-preference phenomenon, suggesting that the content of the responses may also contribute to self-preference bias.

\subsection{Impact of Post-Training Data on Self-Preference}

In this section, we investigate whether fine-tuning two distinct pre-trained models on identical data can help mitigate self-preference bias. Training different models with the same data may encourage the generation of similar responses and align their judgment tendencies. We fine-tune Llama-3.1-8B and Qwen2.5-7B on UltraChat-200k~\citep{ding2023enhancing} using consistent training settings, resulting in Llama-3.1-8B-UltraChat and Qwen2.5-7B-UltraChat. The evaluation results are presented in Figure~\ref{fig:train_data}.

As shown in Figure~\ref{fig:train_data}, fine-tuning different models on the same data helps reduce their self-preference bias. Specifically, the \ourscore scores of Llama-3.1-8B-Instruct and Qwen2.5-7B-Instruct are $10.5\%$ and $2.1\%$, respectively. After fine-tuning with UltraChat-200k, the scores decrease to $2.1\%$ and $1.1\%$. In contrast, for Llama-3.1-8B-Instruct and Qwen2.5-7B-Instruct, which are trained with different data and methods, the \ourscore scores are substantially larger than those observed in their UltraChat-tuned counterparts, reaching $6.1\%$ and $7.6\%$, respectively. Moreover, even after further training on the same dataset, the two models continue to exhibit self-preference bias, suggesting that discrepancies between response generation and evaluation established during pre-training may persist and influence the behavior of downstream fine-tuned models.

\subsection{Attention Analysis}
\begin{figure}[!htbp]
    \centering
    \includegraphics[width=1\linewidth]{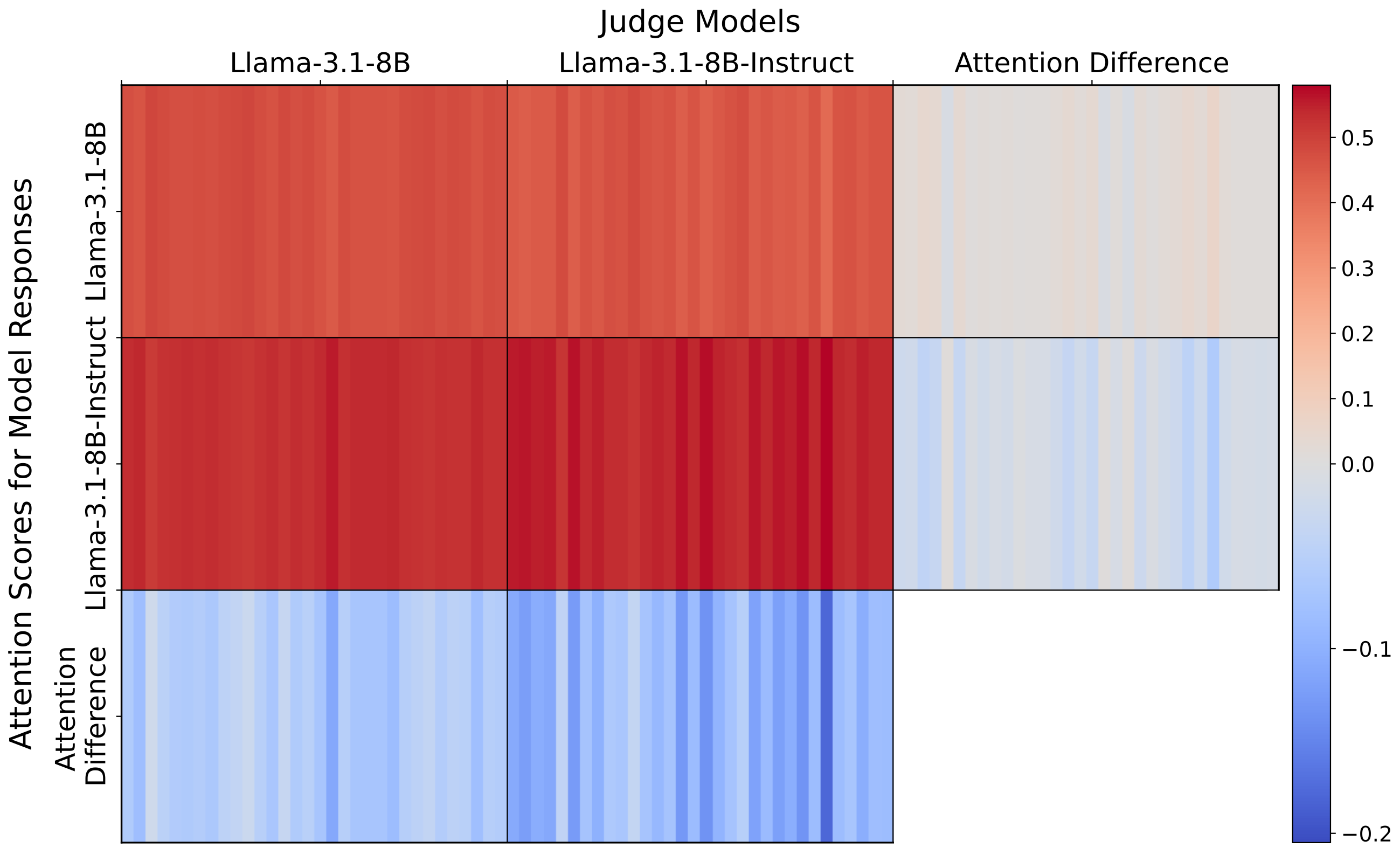}
    \caption{
    Attention scores of each layer in judge models. The scores are averaged over response tokens. The bottom row shows the difference in scores between Llama-3.1-8B and Llama-3.1-8B-Instruct responses for the same judge model. The rightmost column shows the difference in scores assigned by Llama-3.1-8B and Llama-3.1-8B-Instruct (as judges) to the same responses.}
    \label{fig:attn}
    \vspace{-5pt}
\end{figure}

In this section, we analyze self-preference bias from the perspective of attention in LLMs. Specifically, we compare how different judge models allocate attention scores to various responses, aiming to better understand the underlying mechanism of self-preference bias. We use Llama-3.1-8B and Llama-3.1-8B-Instruct as judges and compute the average attention scores over all tokens in the model-generated responses. We then average the attention scores across all test instances and present them for each layer, as shown in Figure~\ref{fig:attn}.

As illustrated in the figure, both judge models assign higher attention scores to the responses generated by Llama-3.1-8B-Instruct compared to those from Llama-3.1-8B (as indicated by the bottom row showing the attention difference). We hypothesize that this is due to the generally higher response quality of Llama-3.1-8B-Instruct, as verified in Figure~\ref{fig:alpaca_main}, which leads to greater attention being paid to its outputs.

Moreover, we also observe that each model tends to assign more attention to its own responses than the other model does. For example, Llama-3.1-8B assigns higher attention to its own responses than Llama-3.1-8B-Instruct does, and vice versa (as indicated by the rightmost column showing the attention difference). This suggests that models naturally allocate more attention to their own responses, contributing to the emergence of self-preference.

\section{Related Work}

\subsection{Large Language Models for Judgment}
LLMs are widely used in judgment tasks such as response ranking~\citep{cui2023ultrafeedback, liu2023alignbench}, reward modeling~\citep{lee2023rlaif, wu2024meta}, and verifying agent task completion~\citep{qin2023toolllm, xia2024evaluating}, driven by their scalability and cost-effectiveness. Leveraging the inherent knowledge and instruction-following abilities of LLMs, researchers can guide these models to perform judgments by directly integrating rules into the prompts~\citep{zheng2023judging, sun2023salmon}. To further refine the judgment capabilities of LLMs in areas such as helpfulness and harmlessness~\citep{bai2022constitutional, wang2023not}, numerous datasets and models have been developed~\citep{lambert2024rewardbench, wang2023pandalm, zhu2023judgelm}, greatly advancing the development and democratization of LLM-based judgment. Another active research area focuses on the meta-evaluation of LLM judges, examining the alignment between LLM judgments and human assessments~\citep{zheng2023judging, dubois2023alpacafarm}, as well as identifying bias in these judges~\citep{koo2023benchmarking, ye2024justice, chen2024humans}.
In this work, we focus on self-preference bias and propose a novel method to more accurately quantify it in LLMs.

\subsection{Bias in Large Language Models}

Extensive studies reveal that LLMs are subject to biases such as length bias~\citep{zheng2023judging, hu2024explaining}, position bias~\citep{zhu2023judgelm, shi2024judging}, and self-preference bias~\citep{ye2024justice, wataoka2024self} in judgment tasks. In this work, we focus on self-preference bias, which refers to the tendency of LLMs to favor their own responses when serving as judges. While several studies have evaluated the presence of self-preference bias in specific models~\citep{ye2024justice, chen2024humans, wang2023large}, a comprehensive analysis across models of different versions, sizes, and reasoning capabilities is still lacking. Although concurrent work~\citep{chen2025llm} conducts large-scale experiments to assess self-preference bias across model families, their focus lies primarily on verifiable tasks such as mathematical reasoning. In contrast, our study centers on open-ended tasks. In addition, several studies have investigated factors related to self-preference bias, such as self-recognition~\citep{panickssery2024llm}, self-enhancement~\citep{xu2024pride}, and preference leakage~\citep{li2025preference}. However, little attention has been given to mitigating this bias. In this work, we make an initial attempt to reduce self-preference bias by exploring two factors: response style and the data used for post-training.

\section{Conclusions}

In this work, we propose the \ourscore score to provide more accurate and reliable measurements of self-preference bias in LLMs. Using this metric, we conduct extensive experiments to evaluate self-preference bias across LLMs of varying versions, sizes, and reasoning abilities. Our further analysis reveals that both the response style and the post-training data of judge models can influence and help alleviate self-preference bias. Additionally, we explore the underlying mechanisms of this bias from an attention-level perspective. Overall, our study underscores the importance of recognizing and addressing self-preference bias when deploying LLMs as judges, and it offers actionable insights into strategies for reducing such bias.

\section*{Limitations}
In this work, we employ GPT-4o-mini, Gemini-1.5-Flash, and DeepSeek-V3 as gold judges to measure the self-preference bias of LLMs. Due to cost constraints, we do not utilize more powerful models, such as GPT-4o or Gemini-1.5-Pro. Using these more capable models could potentially provide more reliable gold-standard judgments, yielding more accurate measurements of self-preference bias. Furthermore, while we mitigate the impact of position bias and length bias through methods like response position swapping and length limitation, other biases, such as authority bias and sentiment bias~\citep{ye2024justice}, may still influence the results. Additionally, this work limits its scope to instruction-following and translation tasks. Further investigation is needed to explore the self-preference bias of LLMs in other tasks, such as agent tasks and dialogue tasks.

\bibliography{latex_tex}

\clearpage
\appendix

\section{Appendix}\label{sec:appendix}

\subsection{Self-Preference of Proprietary Models}\label{appendix:proprietary_models}

\begin{figure*}[!htbp]
  \centering

  \begin{subfigure}{1\textwidth}
    \centering
    \includegraphics[width=1\textwidth]{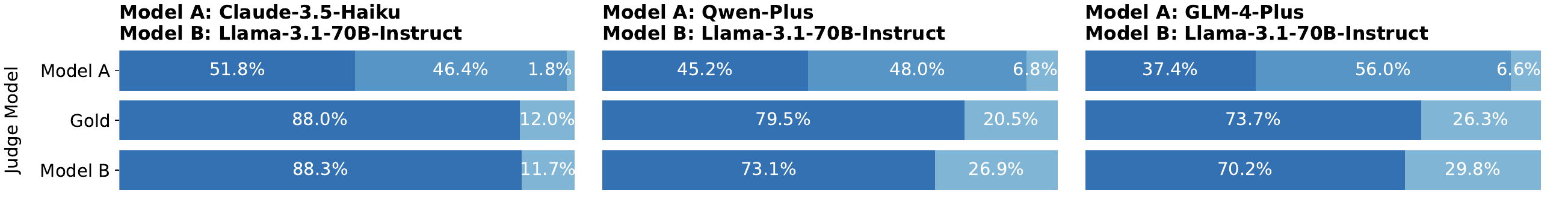}
  \end{subfigure}

  \begin{subfigure}{0.4\textwidth}
    \centering
    \includegraphics[width=\textwidth]{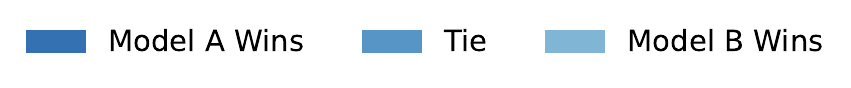}
  \end{subfigure}

  \caption{Judgment results for proprietary models on AlpacaEval.}

  \label{fig:alpaca_close_model}
\end{figure*}

In this section, we attempt to analyze the self-preference bias of proprietary models. We select Claude-3.5-Haiku~\citep{claude2024}, Qwen-Plus~\citep{yang2024qwen2}, and GLM-4-Plus~\citep{glm2024chatglm} for the experiment. Each model is paired with Llama-3.1-70B-Instruct. Since we do not have access to the output probabilities of these models, and preliminary experiments reveal a strong position bias, we classify any test sample where the output tokens (\texttt{A} or \texttt{B}) differ after swapping response positions as a tie in this experiment. The results are shown in Figure~\ref{fig:alpaca_close_model}. Based on the figure, we observe that all three proprietary models exhibit significant position bias, with more than $45\%$ of test samples yielding different judgment results after swapping response positions. When excluding the tied samples, we find that Claude-3.5-Haiku classifies its own responses as superior in $51.8\%/(51.8\% + 1.8\%) = 96.6\%$ of cases, which is higher than the gold judgment of $88.0\%$. This suggests that Claude-3.5-Haiku \textbf{may} exhibit self-preference bias. However, further work is needed to obtain the model's output probabilities to provide more accurate results.

\subsection{Self-Preference on More Datasets}\label{appendix:more_main_results}

\begin{figure*}[!htbp]
  \centering

  \begin{subfigure}{1\textwidth}
    \centering
    \includegraphics[width=1\textwidth]{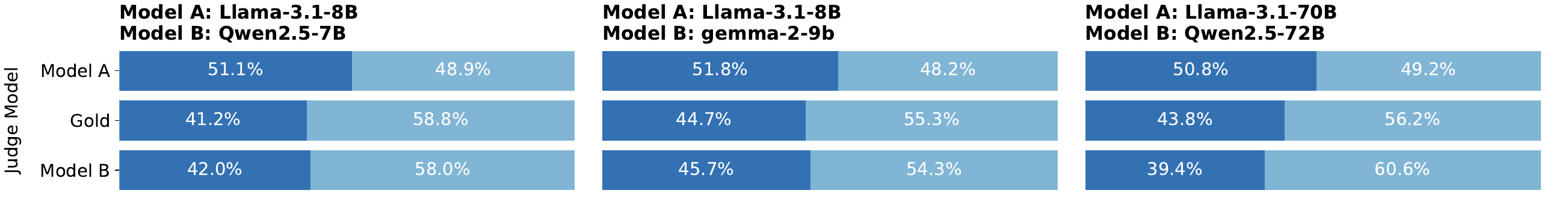}
    \caption{Pairs of pre-trained models of the same size.}
  \end{subfigure}

  \begin{subfigure}{1\textwidth}
    \centering
    \includegraphics[width=1\textwidth]{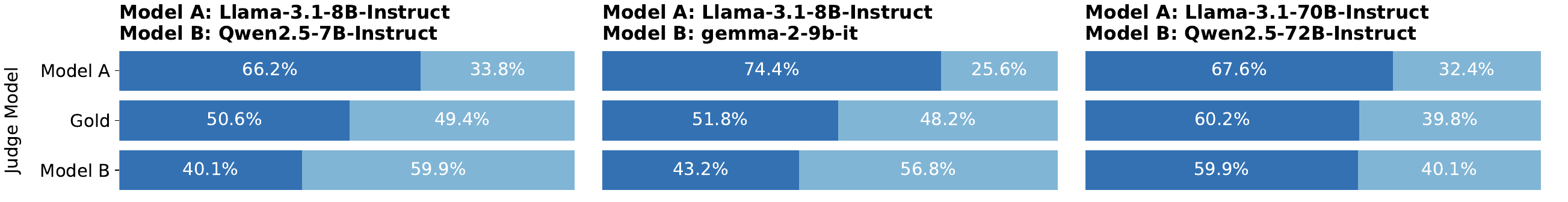}
    \caption{Pairs of post-trained models of the same size.}
  \end{subfigure}

  \begin{subfigure}{1\textwidth}
    \centering
    \includegraphics[width=1\textwidth]{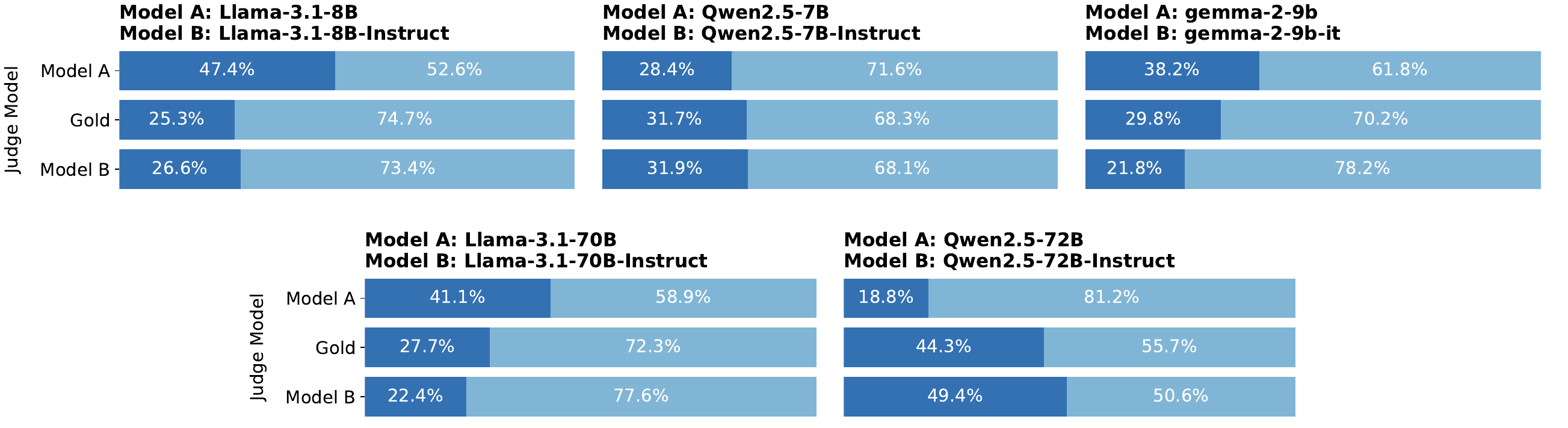}
    \caption{Pairs of pre-trained and post-trained models of the same size.}
  \end{subfigure}

  \begin{subfigure}{1\textwidth}
    \centering
    \includegraphics[width=1\textwidth]{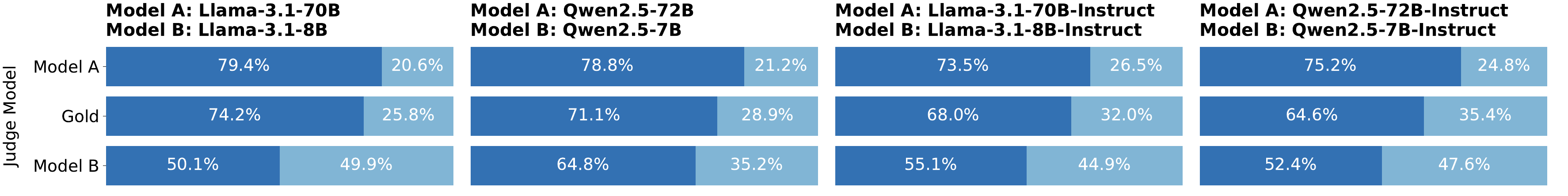}
    \caption{Pairs of models of different sizes.}
  \end{subfigure}

  \begin{subfigure}{0.35\textwidth}
    \centering
    \includegraphics[width=\textwidth]{latex/figure/win_rate/main/legend.pdf}
  \end{subfigure}

  \caption{Judgment results on TruthfulQA.}
  \label{fig:truthfulness_main}
\end{figure*}

\begin{figure*}[!htbp]
  \centering

  \begin{subfigure}{1\textwidth}
    \centering
    \includegraphics[width=1\textwidth]{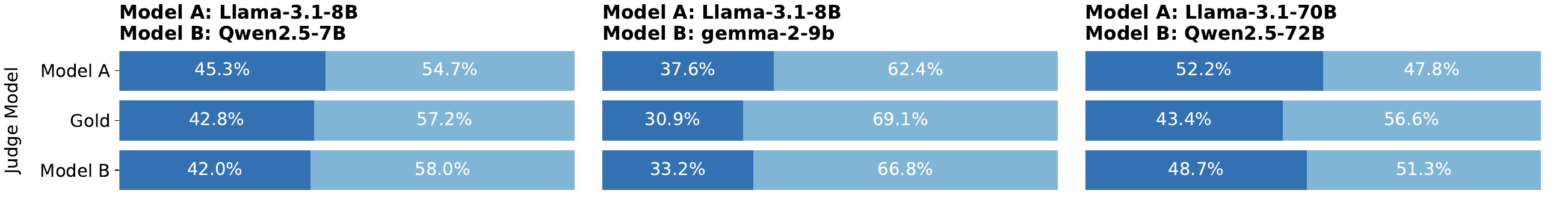}
    \caption{Pairs of pre-trained models of the same size.}
  \end{subfigure}

  \begin{subfigure}{1\textwidth}
    \centering
    \includegraphics[width=1\textwidth]{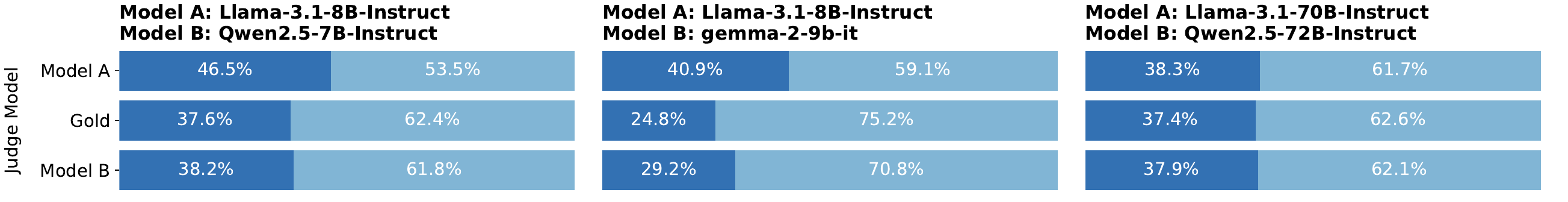}
    \caption{Pairs of post-trained models of the same size.}
  \end{subfigure}

  \begin{subfigure}{1\textwidth}
    \centering
    \includegraphics[width=1\textwidth]{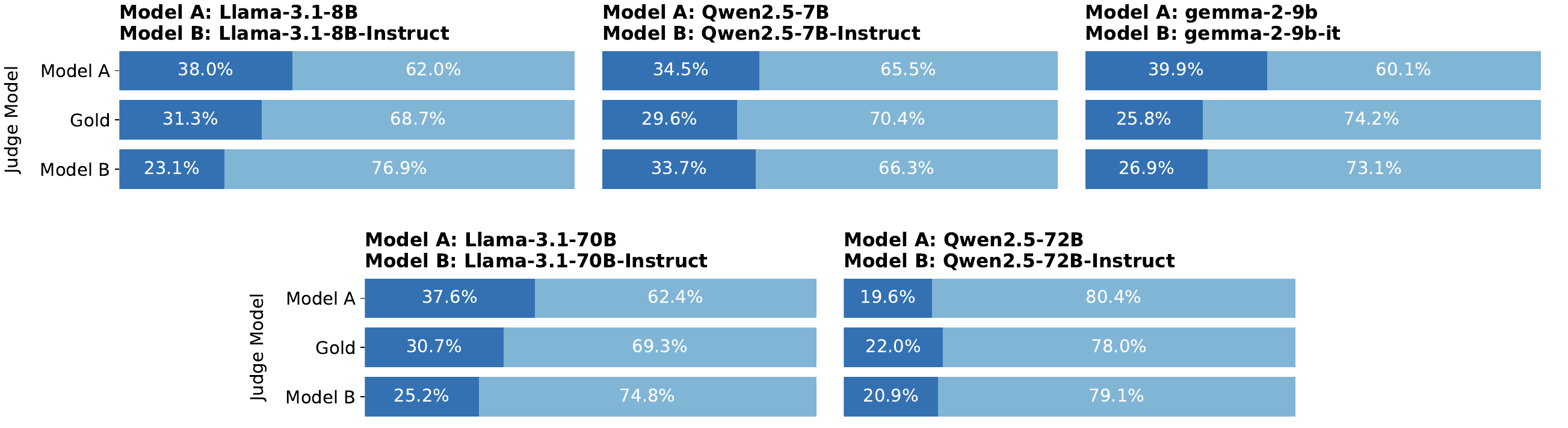}
    \caption{Pairs of pre-trained and post-trained models of the same size.}
  \end{subfigure}

  \begin{subfigure}{1\textwidth}
    \centering
    \includegraphics[width=1\textwidth]{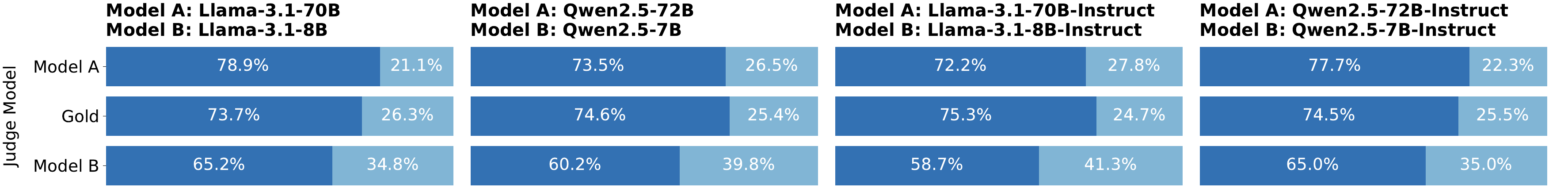}
    \caption{Pairs of models of different sizes.}
  \end{subfigure}

  \begin{subfigure}{0.35\textwidth}
    \centering
    \includegraphics[width=\textwidth]{latex/figure/win_rate/main/legend.pdf}
  \end{subfigure}

  \caption{Judgment results on WMT19 (de-en).}
  \label{fig:translation_main}
\end{figure*}

Figure~\ref{fig:truthfulness_main} and Figure~\ref{fig:translation_main} respectively show the self-preference bias of LLM judges on the TruthfulQA and WMT19 (de-en) datasets. From the figures, we observe similar conclusions to those drawn from AlpacaEval. Specifically, both pre-trained and post-trained models exhibit self-preference bias. For instance, when acting as judges, models like Llama-3.1-8B and Llama-3.1-8B-Instruct tend to give higher scores to their own responses than gold judgments assign to those responses. For example, on the WMT19 (de-en) dataset, when Llama-3.1-8B judges the response pairs of Llama-3.1-8B and Qwen2.5-7B, it exhibits a \ourscore score of  $2.5\%$. Additionally, we observe that large-sized models exhibit less pronounced self-preference bias compared to smaller models. For example, on the TruthfulQA dataset, when large-sized models are paired with small-sized models, the \ourscore scores of the large-sized models tend to be closer to zero than those of the small-sized models.

\subsection{Content Variation in Text Transfer}\label{appendix:content_var}

\begin{figure*}[!htbp]
    \centering
    \begin{minipage}{0.45\textwidth}
        \centering
        \includegraphics[width=\textwidth]{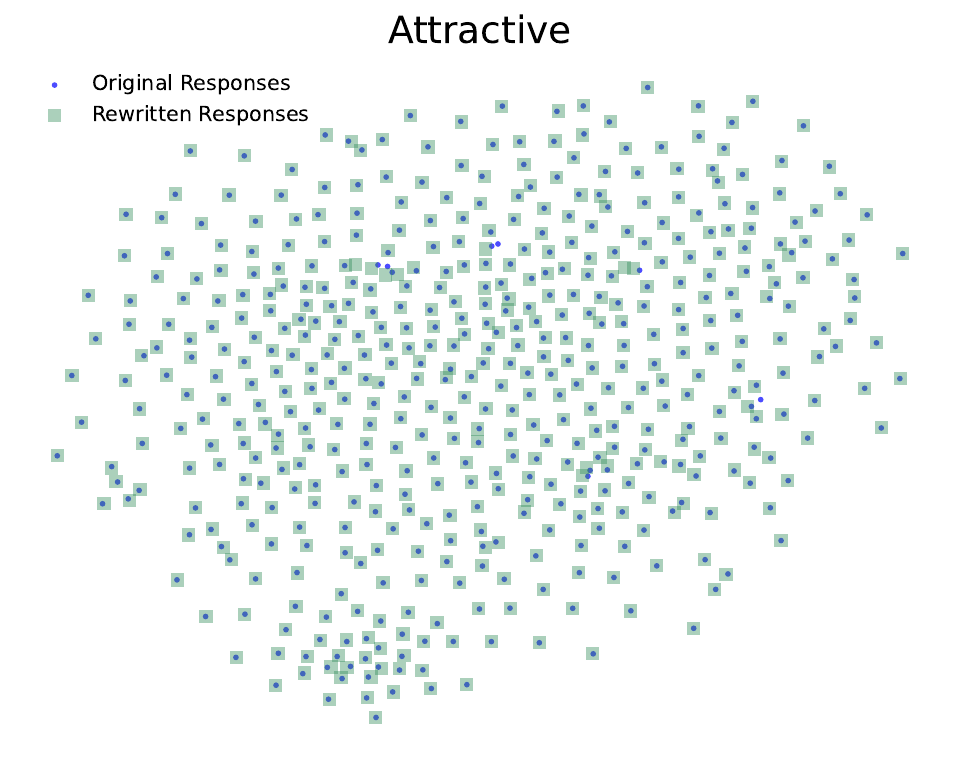} 
    \end{minipage}
    \hspace{0.01\textwidth}
    \begin{minipage}{0.45\textwidth}
        \centering
        \includegraphics[width=\textwidth]{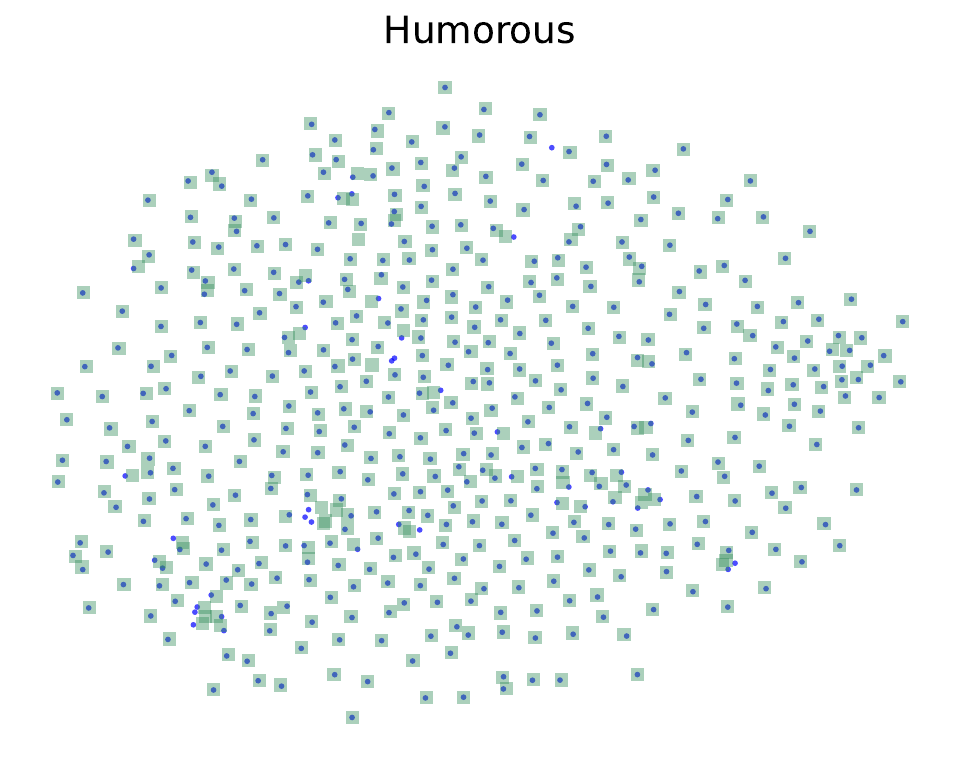} 
    \end{minipage}
    \caption{t-SNE visualizations of the embedding representations of responses before and after rewriting.}
    \label{fig:tsne_plot}
\end{figure*}

To verify that the rewriting approach introduced in Section~\ref{sec:style_transfer} has minimal impact on the semantic content of the text, this section presents an analysis of the representation shifts before and after rewriting. Specifically, we employ gte-multilingual-base~\citep{zhang2024mgte}, a widely-used text representation model, to encode both the original responses generated by Llama-3.1-70B-Instruct and their rewritten counterparts. We use the embedding corresponding to the \texttt{[CLS]} token as the representation of each response. Then, we apply t-SNE~\citep{Laurens2008visualizing} to visualize the changes in representations. The results are shown in Figure~\ref{fig:tsne_plot}. As observed, the representations before and after rewriting exhibit a high degree of overlap, indicating that our rewriting method primarily transfers the style of the responses with minimal impact on their underlying semantics.

\subsection{Consistency Between the Theoretical Bias Estimator and Implementation}\label{appendix:consistency}

In our experimental implementation, for a judge model $A$, we obtain the probabilities assigned to tokens \texttt{A} and \texttt{B} for each individual instruction. The token with the higher probability is selected as the winner. By aggregating the outcomes over all instructions, we compute the win rate, which can be formulated as $\mathbb{E}_x[\mathbb{I}[\sigma(\delta + b_A) > 0.5]]$, where $\mathbb{I}$ is the indicator function. While this procedure produces a binary (0-1) decision for each instruction rather than a continuous probability, it can be viewed as a thresholded approximation to the theoretical quantity $w_A = \mathbb{E}_x[\sigma(\delta + b_A)]$. Specifically, it can be seen as an approximation to sampling from a Bernoulli distribution with success probability $\sigma(\delta + b_A)$. The same applies to gold judge models. The approximation error is small when the underlying probabilities are well-separated (i.e., close to 0 or 1). This justifies the empirical procedure as a practical surrogate to the theoretical self-preference bias formulation.

\begin{table}[!htbp]
    \scriptsize
    \centering
    \setlength{\tabcolsep}{3pt}
    \begin{tabular}{clrr}
        \toprule
        \multirow{2}{*}{Model} & \multirow{2}{*}{Paired Response} & \multicolumn{2}{c}{Win Rate} \\
        \cmidrule(lr){3-4}
        &  & Zero-shot & Few-shot \\
        \midrule
        \multirow{2}{*}{\ \ Llama-3.1-8B-Instruct} & Llama-3.1-8B & 82.0\% & 78.0\% \\
        & Qwen2.5-7B-Instruct & 54.0\% & 53.2\% \\
        \midrule
        \multirow{2}{*}{\quad Llama-3.1-70B-Instruct} & Llama-3.1-70B & 83.5\% & 84.9\% \\
        & Qwen2.5-72B-Instruct & 50.0\% & 48.0\% \\
        \midrule
        \multirow{2}{*}{Qwen2.5-7B-Instruct} & Qwen2.5-7B & 68.1\% & 69.8\% \\
        & Llama-3.1-8B-Instruct & 59.7\% & 60.6\% \\
        \bottomrule
    \end{tabular}
    \caption{Comparison of post-trained models judgments to their responses under zero-shot and few-shot settings.}

    \label{tab:few_shot_setting}
\end{table}

\subsection{Few-shot Setting Analysis}
To guide pre-trained models in making judgments, we leverage their few-shot learning ability and prepend examples to each input. For post-trained models, due to their strong instruction-following ability, we prompt them to make judgments in a zero-shot setting. To investigate the differences in judgment between zero-shot and few-shot settings for post-trained models, we conduct judgment experiments under the few-shot setting. The results are shown in Table~\ref{tab:few_shot_setting}. From the table, we observe that the judgment results of the post-trained model in the zero-shot and few-shot settings are similar, indicating that the post-trained model is capable of generating appropriate judgments in the zero-shot setting, which validates the reasonableness of our experimental setup.

\subsection{Prompt}\label{appendix:prompt}
We present the prompts used for response generation in Table~\ref{table:alpaca_response}, Table~\ref{table:truthfulness_response}, and Table~\ref{table:translation_response}, and the prompts used for response judgment in Table~\ref{table:alpaca_judgment}, Table~\ref{table:truthfulness_judgment}, and Table~\ref{table:translaiton_judgment}.

\begin{table*}[!htbp]
    \small
    \centering
    \begin{tabular}{p{1.98\columnwidth}}
    \toprule
    \textit{Post-trained Models}: \\
You are a helpful assistant. Please help me answer the query below. Keep your answer within 200 words. \\
\\
\# Query \\
\{query\} \\
\\
Now, please provide your answer to this query. \\
\# Answer \\
    \midrule
    \textit{Pre-trained Models}: \\
\# Instruction \\
Below is a list of conversations between a human and an AI assistant (you). \\
As an AI assistant, you will engage in conversations with users, responding to their queries which are presented under the heading "\# Query". \\
Your responses should be entered under the heading "\# Answer". \\
You excel in a wide range of tasks including, but not limited to, providing general information, conducting reasoning, engaging in role-play, creative writing, planning, and solving mathematical and coding problems. \\
Your responses should be well-structured, comprehensive, and aim to thoroughly address the user's query or problem at hand.
**Please keep your answer within 200 words.** \\
\\
\# Query \\
\{example 1 query\} \\
\# Answer \\
\{example 1 answer\} \\
\\
\# Query \\
\{example 2 query\} \\
\# Answer \\
\{example 2 answer\} \\
\\
\# Query \\
\{example 3 query\} \\
\# Answer \\
\{example 3 answer\} \\
\\
\# Query \\
\{test query\} \\
\# Answer \\
    \bottomrule
    \end{tabular}
    \caption{Prompt used for response generation on AlpacaEval. The prompt for pre-trained models is derived from \citet{lin2023unlocking}.}
    \label{table:alpaca_response}
\end{table*}

\begin{table*}[!htbp]
    \small
    \centering
    \begin{tabular}{p{1.98\columnwidth}}
    \toprule
    \textit{Post-trained Models}: \\
You are a truthful assistant. Please help me answer the query below truthfully, remaining faithful to factual knowledge and the given context. Avoid making up any untrue facts. \\
\\
\# Query \\
\{query\} \\
\\
Now, please provide your answer to the query. Keep your answer within 200 words. \\
\# Answer \\
    \midrule
    \textit{Pre-trained Models}: \\
\# Instruction \\
As an AI assistant, you will engage in conversations with users, responding to their queries which are presented under the heading "\# Query". \\
Your responses should be entered under the heading "\# Answer". \\
**You should respond to queries truthfully, remaining faithful to factual knowledge and the given context. Avoid making up any untrue facts.** \\
Keep your answers within 200 words. \\
\\
\# Query \\
\{example 1 query\} \\
\# Answer \\
\{example 1 answer\} \\
\\
\# Query \\
\{example 2 query\} \\
\# Answer \\
\{example 2 answer\} \\
\\
\# Query \\
\{test query\} \\
\# Answer \\
    \bottomrule
    \end{tabular}
    \caption{Prompt used for response generation on TruthfulQA.}
    \label{table:truthfulness_response}
\end{table*}

\begin{table*}[!htbp]
    \small
    \centering
    \begin{tabular}{p{1.98\columnwidth}}
    \toprule
    \textit{Post-trained Models}: \\
You are an excellent translator, and you specialize in translating German into English. Given a piece of German text, please help translate it into English. \\
Here is the given German text. \\
\# German \\
\{german\} \\
\\
Now, please translate the German text into English. You only need to provide the English translation, with no other text. \\
\# English \\
    \midrule
    \textit{Pre-trained Models}: \\
\# Instruction \\
You are an excellent translator, and you specialize in translating German into English. **Given a piece of German text, please translate it into English.** \\
The German texts are under "\# German", and the corresponding English translations are under "\# English". \\
\\
\# German \\
\{example 1 german\} \\
\# English \\
\{example 1 english\} \\
\\
\# German \\
\{example 2 german\} \\
\# English \\
\{example 2 english\} \\
\\
\# German \\
\{test german\} \\
\# English \\
    \bottomrule
    \end{tabular}
    \caption{Prompt used for response generation on WMT19 (de-en).}
    \label{table:translation_response}
\end{table*}

\begin{table*}[!htbp]
    \small
    \centering
    \begin{tabular}{p{1.98\columnwidth}}
    \toprule
    \textit{Post-trained Models}: \\
You are a helpful assistant tasked with evaluating the quality of different responses to a given query. For each query, you will receive two independent responses. Please judge which response is better. \\
\\
Here is the given query. \\
\# Query \\
\{query\} \\
\\
Here are two independent responses (Response A and Response B) to the query. \\
\# Responses \\
<Response A>\{response a\}</Response A> \\
<Response B>\{response b\}</Response B> \\
\\
Now, please judge which response is better. You only need to output A or B, with no other text. \\
Your output should be A if Response A is better or B if Response B is better. \\
Please remember that your response should start with either A or B \\

    \midrule
    
    \textit{Pre-trained Models}: \\
\# Instruction \\
You are a helpful assistant tasked with evaluating the quality of different responses to a query. You will receive multiple queries, each accompanied by two independent responses. For each query, you should judge which response you think is better. \\
Queries are placed under "\# Query". The two responses for each query are listed under "\# Responses" as "Response A" and "Response B". Your outputs should be placed under "\# Judgment". \\
You only need to output A or B, with no other text. \\
\\
\# Query \\
\{example 1 query\} \\
\# Responses \\
<Response A>\{example 1 response a\}</Response A> \\
<Response B>\{example 1 response b\}</Response B> \\
\# Judgment \\
\{example 1 judgment\} \\
\\
\# Query \\
\{example 2 query\} \\
\# Responses \\
<Response A>\{example 2 response a\}</Response A> \\
<Response B>\{example 2 response b\}</Response B> \\
\# Judgment \\
\{example 2 judgment\} \\
\\
\# Query \\
\{test query\} \\
\# Responses \\
<Response A>\{test response a\}</Response A> \\
<Response B>\{test response b\}</Response B> \\
\# Judgment \\
    \bottomrule
    \end{tabular}
    \caption{Prompt used for judgment generation on AlpacaEval.}
    \label{table:alpaca_judgment}
\end{table*}

\begin{table*}[!htbp]
    \small
    \centering
    \begin{tabular}{p{1.98\columnwidth}}
    \toprule
    \textit{Post-trained Models}: \\
You are a knowledgeable assistant. Given a query and two corresponding responses, your task is to determine which response is more truthful, ensuring it is less misleading, free from fabricated content, factual inaccuracies, and logical errors. \\
\\
Here is the query. \\
\# Query \\
\{query\} \\
\\
Here are two responses (Response A and Response B) to the query. \\
\# Responses \\
<Response A>\{response a\}</Response A> \\
<Response B>\{response b\}</Response B> \\
\\
Now, please judge which response is more truthful. You only need to output A or B, with no other text.
Please remember that your response should start with either A or B. \\

    \midrule
    
    \textit{Pre-trained Models}: \\
\# Instruction \\
You are a knowledgeable assistant. Given a query and two corresponding responses, your task is to determine which response is **more truthful**, ensuring it is less misleading, free from fabricated content, factual inaccuracies, and logical errors. \\
Queries are placed under "\# Query". The two responses for each query are listed under "\# Responses" as "Response A" and "Response B". Your outputs should be placed under "\# Judgment". \\
You only need to output A or B, with no other text. \\
\\
\# Query \\
\{example 1 query\} \\
\# Responses \\
<Response A>\{example 1 response a\}</Response A> \\
<Response B>\{example 1 response b\}</Response B> \\
\# Judgment \\
\{example 1 judgment\} \\
\\
\# Query \\
\{example 2 query\} \\
\# Responses \\
<Response A>\{example 2 response a\}</Response A> \\
<Response B>\{example 2 response b\}</Response B> \\
\# Judgment \\
\{example 2 judgment\} \\
\\
\# Query \\
\{test query\} \\
\# Responses \\
<Response A>\{test response a\}</Response A> \\
<Response B>\{test response b\}</Response B> \\
\# Judgment \\
    \bottomrule
    \end{tabular}
    \caption{Prompt used for judgment generation on TruthfulQA.}
    \label{table:truthfulness_judgment}
\end{table*}

\begin{table*}[!htbp]
    \small
    \centering
    \begin{tabular}{p{1.98\columnwidth}}
    \toprule
    \textit{Post-trained Models}: \\
You are a helpful assistant tasked with evaluating the quality of two different English translations of the same German text. For each German text, you will receive two independent English translations. Please judge which English translation is better. \\
\\
Here is the German text. \\
\# German \\
\{german\} \\
\\
Here are two independent English translations (English A and English B) for the German text. \\
\# English \\
<English A>\{english a\}</English A> \\
<English B>\{english b\}</English B> \\
\\
Now, please judge which English translation is better. You only need to output A or B, with no other text. Please remember that your response should start with either A or B \\

    \midrule
    
    \textit{Pre-trained Models}: \\
\# Instruction \\
You are a helpful assistant tasked with evaluating the quality of two different English translations of the same German text. For each German text, you will receive two independent English translations. Please judge which English translation is better. \\
The German texts are under "\# German". The two independent English translations for each German text are under "\# English", labeled as "English A" and "English B", respectively. Your outputs should be placed under "\# Judgment". \\
You only need to output A or B, with no other text. \\
\\
\# German \\
\{example 1 german\} \\
\# English \\
<English A>\{example 1 english a\}</English A> \\
<English B>\{example 1 english b\}</English B> \\
\# Judgment \\
\{example 1 judgment\} \\
\\
\# German \\
\{example 2 german\} \\
\# English \\
<English A>\{example 2 english a\}</English A> \\
<English B>\{example 2 english b\}</English B> \\
\# Judgment \\
\{example 2 judgment\} \\
\\
\# German \\
\{test german\} \\
\# English \\
<English A>\{test english a\}</English A> \\
<English B>\{test english b\}</English B> \\
\# Judgment \\
    \bottomrule
    \end{tabular}
    \caption{Prompt used for judgment generation on WMT19 (de-en).}
    \label{table:translaiton_judgment}
\end{table*}


    

\end{document}